\documentclass[runningheads]{llncs}
\usepackage{graphicx}
\usepackage{amsmath}
\usepackage{amssymb}
\usepackage{multirow}
\usepackage{subfigure}
\usepackage{graphicx}
\usepackage{pgfplots}

\DeclareMathOperator*{\argmax}{argmax}
\begin{document}
\title{Multi-Branch Siamese Networks with Online Selection for Object Tracking}
%
%
 \author{Zhenxi Li\inst{1} \and Guillaume-Alexandre Bilodeau\inst{1} \and Wassim Bouachir\inst{2}}
 \authorrunning{Z. Li et al.}
%
 \institute{\textsuperscript{1}LITIV lab, Polytechnique Montreal\\
 \email{\{zhenxi.li, guillaume-alexandre.bilodeau\}@polymtl.ca}\\
 \textsuperscript{2}TELUQ University\\
 \email{wassim.bouachir@teluq.ca}}
\maketitle              
\begin{abstract}
In this paper, we propose a robust object tracking algorithm based on a branch selection mechanism to choose the most efficient object representations from multi-branch siamese networks. While most deep learning trackers use a single CNN for target representation, the proposed Multi-Branch Siamese Tracker (MBST) employs multiple branches of CNNs pre-trained for different tasks, and used for various target representations in our tracking method. With our branch selection mechanism, the appropriate CNN branch is selected depending on the target characteristics in an online manner. By using the most adequate target representation with respect to the tracked object, our method achieves real-time tracking, while obtaining improved performance compared to standard Siamese network trackers on object tracking benchmarks.

\end{abstract}
\section{Introduction}
Model-free visual object tracking is one of the most fundamental problems in computer vision. Given the object of interest marked in the first video frame, the objective is to localize the target in subsequent frames, despite object motion, changes in viewpoint, lighting variation, among other disturbing factors. One of the most challenging difficulties with model-free tracking is the lack of prior knowledge on the target object appearance. Since any arbitrary object may be tracked, it is impossible to train a fully specialized tracker.

Recently, convolutional neural networks (CNNs) have demonstrated strong power in learning feature representations. To fully exploit the representation power of CNNs in visual tracking, it is desirable to train them on large datasets specialized for visual tracking, and covering a wide range of variations in the combination of target and background. However, it is truly challenging to learn a unified representation based on videos that have completely different characteristics. Some trackers~\cite{goturn} train regression networks for tracking in an entirely offline manner. Other works~\cite{siamfc,cfnet,sasiam} propose to train deep CNNs to address the general similarity learning problem in an offline phase and evaluate the similarity online during tracking. However, since these works have no online adaptation, the representations they learned offline are general but not always discriminative.

Rather than applying a single fixed network for feature extraction, we propose to use multiple network branches with an online branch selection mechanism. It is well known that different networks designed and trained for different tasks have diverse feature representations. With the online branch selection mechanism, our tracker dynamically selects the most efficient and robust branch for target representation, even if the target appearance changes. Our goal is to improve the generalization capability with multiple networks.

The main contributions of our work are summarized as follows. First, we propose a multi-branch framework based on a siamese network for object tracking. The proposed architecture is designed to extract appearance representation robust against target variations and changing contrast with background scene elements. Second, to make the full use of the different branches, we propose an effective and generic branch selection mechanism to dynamically select branches according to their discriminative power. Third, on the basis of multiple branches and branch selection mechanism, we present a novel deep learning tracker achieving real-time and improved tracking performance. Our extensive experiments compare the proposed Multi-Branch Siamese Tracker (MBST) with state-of-the-art trackers on OTB benchmarks~\cite{otb13,otb15}.

\section{Related Work}
\textbf{Siamese Network Based Trackers.}
Object tracking can be addressed using similarity learning. By learning a deep embedding function, we can evaluate the similarity between an exemplar image patch and a candidate patch in a search region. These procedures allow to track the target to the location that obtains the highest similarity score. Inspired by this idea, the pioneering work of SiamFC~\cite{siamfc} proposed a fully-convolutional Siamese Network in which the similarity learning with deep CNNs is addressed using a Siamese architecture. Since this approach does not need online training, it can easily achieve real-time tracking. Due to the robustness and real-time performance of the SiamFC~\cite{siamfc} approach, several subsequent works proceeded along this direction to address the tracking problem. In this context, EAST~\cite{east} employs an early-stopping agent to speed up tracking where easy frames are processed with cheap features, while challenging frames are processed with deep features. CFNet~\cite{cfnet} incorporates a Correlation Filter into a shallow siamese network, which can speed up tracking without accuracy drop comparing to a deep Siamese network. TRACA~\cite{traca} applies context-aware feature compression before tracking to achieve high tracking performance. SA-Siam~\cite{sasiam} utilizes the combination of semantic features and appearance features to improve generalization capability. In our work, we use the Siamese Network as embedding function to extract feature representations. All branches use the Siamese architecture to apply identical transformation on target patch and search region.

\textbf{Multi-Branch Tracking Frameworks.}
The diversity of target representation from a single fixed network is limited. The learned features may not be discriminative in all tracking situations. There are many works using diverse features with context-aware or domain-aware scheme.

TRACA~\cite{traca} is a multi-branch tracker, which utilizes multiple expert auto-encoders to robustly compress raw deep convolutional features. Since each of expert auto-encoders is trained according to a different context, it performs context-dependent compression. MDNet~\cite{mdnet} is composed of shared layers and multiple branches of domain-specific layers. BranchOut~\cite{branchout} employs a CNN for target representation, with a common convolutional layers and multiple branches of fully connected layers. It allows different number of layers in each branch to maintain variable abstraction levels of target appearances. 

A common insight of these multi-branch trackers is the possibility to make a robust tracker by utilizing different feature representations. Our method shares some insights and design principles with other multi-branch trackers. Our network architecture is composed of multiple branches separately trained offline and focusing on different types of CNN features. In addition, we use an AlexNet~\cite{alexnet} branch in our framework that is designed and pretrained for image classification. In our multi-branch frameworks, the combination of branches trained in different scenarios ensures a better use of diverse feature representations. 

\textbf{Online Branch Selection.}
Different models produce various feature maps on different tracked targets in different scales, rotations, illumination and other factors. Using all features available for a single object tracking is neither efficient nor effective. BranchOut~\cite{branchout} selects a subset of branches randomly for model update to diversify learned target appearance models. MDNet~\cite{mdnet} learns domain-independent representations from pretraining, and identifies branches through online learning. 

In our online branch selection mechanism, we analyse the feature representation of each branch to select the most robust branch at every $T$ frames. This allows us to use diverse feature representations and to handle various challenges in the object tracking problem more efficiently.

\section{Multi-Branch Siamese Tracker}
We propose a multi-branch siamese network for tracking. Given that different neural network models produce diverse feature representations, we use many of them as branches in our tracker to produce diverse feature representations and select the most robust branch with our online branch selection mechanism.

\begin{figure}
\centering
\includegraphics[width=0.8\linewidth]{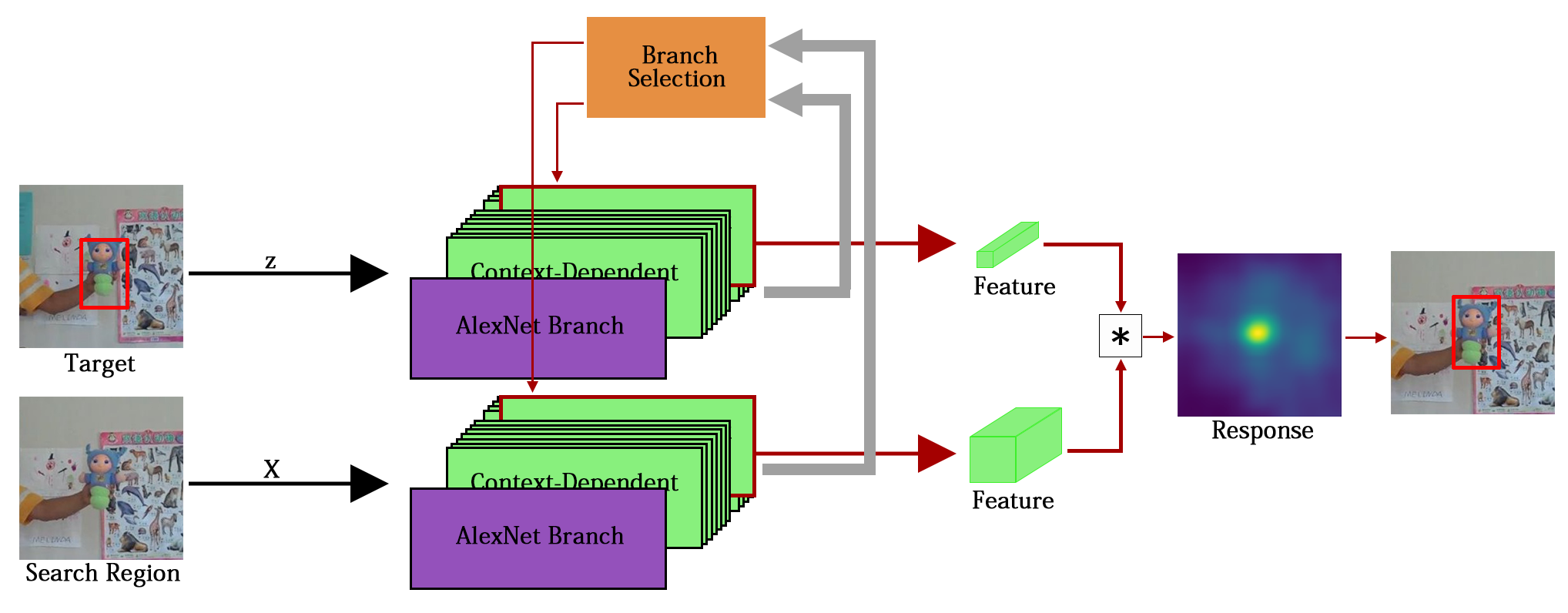}
\caption{The architecture of our MBST tracker. Context-dependent branches are indicated by green blocks and AlexNet branch is indicated by purple blocks.
}\label{fig1}
\end{figure}

\subsection{Network Architecture}
Using multiple target representations is shown to be beneficial for object tracking~\cite{sasiam,tcnn}, as different CNNs can provide various feature representations. In our work, we ensemble $N_{e}$ siamese networks including $N_{s}$ context-dependent branches and one AlexNet branch as $N_{e} = N_{s}+1$. The context-dependent branches have the same structure as SiamFC~\cite{siamfc} and the AlexNet branch has the same structure as AlexNet~\cite{alexnet}. Each branch of the tracker is a siamese network applying identical transformation $\varphi_{i}$ to both inputs and combining their representation by a cross-correlation layer. The architecture of the proposed tracker is illustrated in Fig.~\ref{fig1}. 

The input consists of a target patch cropped from the first video frame and another patch containing the search region in the current frame. The target patch $z$ has a size of $W_{z}\times H_{z}\times 3$, corresponding to the width, height and color channels of the image patch. The search region $X$ has a size of $W_{X}\times H_{X}\times 3$ ($W_{z}<W_{X}$ and $H_{z}<H_{X}$), representing also the width, height and color channels of the search region. $X$ can be considered as a collection of candidate patches $x$ in the search region with the same dimension as $z$. 

From what we observed, there are two strategies to improve the discriminative ability of the tracking networks. The first one is training the network in different contexts, while the second one is to use multiple networks designed and trained for different tasks. In our approach, we utilize context-dependent branches pretrained in different contexts in addition to another branch pretrained for image classification task to improve our tracking performance. We note that more branches could be added with other pre-trained networks at the cost of slower performances.

\textbf{Context-dependent branches:} We use $N_{c}$ context-dependent branches and one general branch as $N_{s}=N_{c}+1$. All these branches have the same architecture as the SiamFC network~\cite{siamfc}. Context-dependent branches are trained in three steps. Firstly, we train the basic siamese network on the ILSVRC-2015~\cite{imagenet} video dataset (henceforth ImageNet), including 4,000 video sequences and around 1.3 million frames containing about 2 million tracked objects. We keep the basic siamese network as the general branch. Then, we perform contextual clustering on the low level feature map from the ImageNet Video dataset to find $N_{c}$ ($N_{c}=10$) context-dependent clusters. Finally, we use the $N_{c}$ clusters to train $N_{c}$ context-dependent branches initialized by the basic siamese network. These branches take $(z,X)$ as input and extract their feature maps. Then, using a cross correlation layer we combine their feature maps to get a response map. The response map of context-dependent branches is calculated as:

\begin{figure}
\centering
\includegraphics[width=0.8\linewidth]{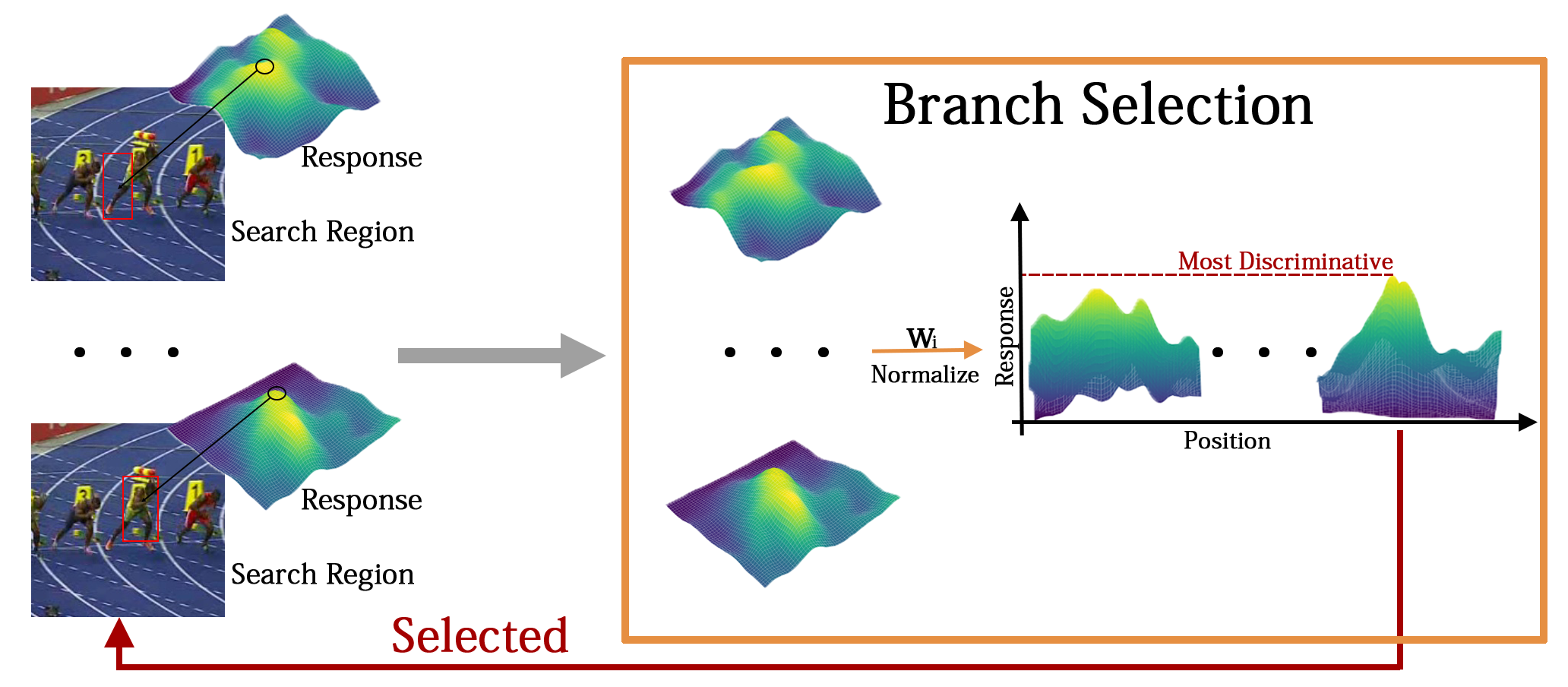}
\caption{Online branch Selection mechanism and response map example.}\label{fig2}
\end{figure}
\begin{equation}
h_{s_{i}}(z,X)=corr(f_{s_{i}}(z),f_{s_{i}}(X)), 
\end{equation}
where $s_{i}$ indicates the contextual index including the general branch ($i=0$), $f(\cdot)$ denotes features generated by the network.

\textbf{The AlexNet branch:} We use AlexNet~\cite{alexnet} pretrained on the image classification task as a branch with a network trained for a different task. Small modifications are made on the stride to ensure that the output response map has the same dimension as other branches. Since AlexNet is trained for image classification and the deeper layers encode more semantic information of targets, target representations from this branch are more robust to significant appearance variations. The network output corresponds to $(z,X)$ as input, while the generated features are denoted as $f_a(\cdot)$. The response map is expressed as:
\begin{equation}
h_{a}(z,X)=corr(f_{a}(z),f_{a}(X)).
\end{equation}

In our implementation, MBST is composed of context-dependent branches and AlexNet branch. The output of each branch is a response map indicating the similarity between target $z$ and candidate patch $x$ within the search region $X$. The branch selection mechanism compares the maps from each branch to select the most discriminative one. The corresponding branch is then used for $T-1$ frames.

\subsection{Online Branch Selection Mechanism}
Different branches trained in different scenarios can be used to diversify the target representation. To ensure the optimal exploitation of the diverse representations from our branches, we designed a branch selection mechanism to monitor the tracking output and automatically select the most discriminative branch as illustrated in Fig.~\ref{fig2}.

Given the input image pair, each branch applies identical transformation to both inputs and calculates the response map $h$ using a cross-correlation layer. Since the ranges of feature values from different branches are different, we apply response weights $w_{i}$ on response map of each branches to normalize their range difference. The discriminative power is then measured based on the weighted response maps from all branches. The heuristic approach we used to measure the discriminative power of branches is formulated as:
\begin{equation}
R(w_{i}h_{B_{i}}) = w_{i}(P(h_{B_{i}})-M(h_{B_{i}})),
\end{equation} 
where $h_{B_{i}}$ is the response map for each branch $B_{i}$, $P_{B_{i}}$ is the peak value of the response map $h_{B_{i}}$, and $M_{h_{B_{i}}}$ is the minimum value of the response map $h_{B_{i}}$. 

The objective function of our branch selection mechanism can be written as:
\begin{equation}
B^{*}=\argmax_{B_{i}}R(w_{i}h_{B_{i}}),
\end{equation} 
where $B^{*}$ is the selected branch to transform inputs.

\section{Experiments}
The first aim of our experiments is to investigate the effect of incorporating multiple feature representations with an online branch selection mechanism. For this purpose, we performed ablation analysis on our framework. We then compare our method with state-of-the-art trackers. The experimental results demonstrate that our method achieves improved performance with respect to the basic SiamFC tracker~\cite{siamfc}.

\subsection{Implementation Details}
\textbf{Network structure:} The context-dependent branches have exactly the same structure as the SiamFC network~\cite{siamfc}. For the AlexNet branch, we use AlexNet~\cite{alexnet} pretrained on ImageNet dataset~\cite{imagenet} with a small modification to ensure that the output response map has the same dimension as other branches, which is $17\times 17$. Other branches could also be used based on other network architectures.

\textbf{Data Dimensions:} In our experiment, the target image patch $z$ has a dimension of $127\times 127\times 3$, and the search region $X$ has a dimension of $255\times 255\times 3$. But since all branches are fully convolution layers, they can also be adapted to any other dimension easily. The embedding output for $z$ and $X$ has a dimension of $6\times 6\times 256$ and $22 \times 22 \times 256$ respectively.

\textbf{Training:} We use the ImageNet dataset~\cite{imagenet} for training and only consider color images. For simplicity, we randomly pick a pair of images, we crop $z$ in the center and $X$ in the center of another image. Images are scaled such that the bounding box, plus an added margin for context, has a fixed area. The basic siamese branch is trained for 50 epochs with an initial learning rate of 0.01. The learning rate decays after every epoch with a decay factor $\delta$ of 0.869. The context-dependent branches are fine-tuned based on the parameters of the general branch with a learning rate 0.00001 for 10 epochs. For the AlexNet branch, we directly use AlexNet~\cite{alexnet} pretrained on ImageNet dataset~\cite{imagenet}.

Our experiments are performed on a PC with a Intel i7-3770 3.40 GHz CPU and a Nvidia Titan X GPU. We evaluated our results using the Python implementation of the OTB toolkit. The average testing speed of MBST is 17 fps.

\begin{table}
\centering
\caption{Ablation study of MBST on OTB benchmarks. Various combinations of general siamese branch, context-dependent branches and AlexNet branch are evaluated.}\label{tab1}
  \begin{tabular}{l l l|l l|l l|l l| l}
    \hline
     & & &
      \multicolumn{2}{c|}{OTB-2013} &
      \multicolumn{2}{c|}{OTB-50} &
      \multicolumn{2}{c|}{OTB-100} & \\
    General & Context & AlexNet & AUC & Prec. & AUC & Prec. & AUC & Prec. & FPS\\
    \hline
    \checkmark & & & 0.600 & 0.791 & 0.519 & 0.698 & 0.585 & 0.766 & \textbf{65.0}\\
    
    & \checkmark & & 0.601 & 0.798 & 0.523 & 0.707 & 0.584 & 0.768 & 18.6 \\
    
    & & \checkmark & 0.581 & 0.761 & 0.501 & 0.678 & 0.560 & 0.741 & 63.6 \\
    
    \checkmark & \checkmark & & 0.594 & 0.784 & 0.535 & 0.721 & 0.587 & 0.770 & 16.9 \\
    
    \checkmark & & \checkmark & 0.605 & 0.796 & 0.536 & 0.718 & 0.599 & 0.783 & 42.9 \\
    
    & \checkmark & \checkmark & 0.616 & 0.811 & 0.570 & 0.767 & 0.614 & 0.806 & 16.9 \\
    
    \checkmark & \checkmark & \checkmark & \textbf{0.620} & \textbf{0.816} & \textbf{0.573} & \textbf{0.773} & \textbf{0.617} & \textbf{0.811} & 16.9\\
    \hline
  \end{tabular}
\end{table}

\textbf{Hyperparameters:} The weights $w_{i}$ for context-dependent branches have the same value of 1.0. For AlexNet branch, we perform a grid search from 8.0 to 12.0 with step 0.5. Evaluation suggests that the best performance is achieved when $w_{i}$ is 10.5. This value is thus used for all the test sequences. In order to handle scale variations, we rescale the inputs into three different resolutions.

\subsection{Dataset and Evaluation Metrics}
\textbf{OTB:} We evaluate the proposed tracker on the OTB benchmarks~\cite{otb13,otb15} with eleven interference attributes for the video sequences. The OTB benchmark uses the precision and success rate for quantitative analysis. For the precision plot, we calculate the average Euclidean distance between the center locations of the tracked targets and the manually labeled ground truth. Then the average center location error over all the frames of one sequences is used to summarize the overall performance. As the representative precision score for each tracker, we use the score for the threshold of 20 pixels. For the success plot, we compute the IoU (intersection over union) between the tracked and ground truth bounding boxes. A success plot is obtained by evaluating the success rate at different IoU thresholds. The area-under-curve (AUC) of the success plot is reported.

\subsection{Ablation Analysis}
To verify the contribution of each branch and the online branch selection mechanism of our algorithm, we implemented several variations of our approach and evaluated them on the OTB benchmarks.

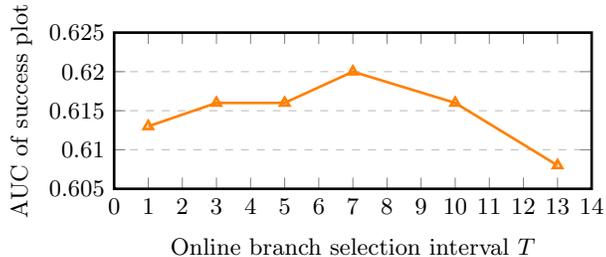
\begin{figure}
\centering
    \begin{tikzpicture}
    \begin{axis}[
    line width=0.35mm,
    width=0.65\textwidth,
    height=0.3\textwidth,
    xlabel={Online branch selection interval $T$},
    ylabel={AUC of success plot},
    xmin=0, xmax=14,
    ymin=0.605, ymax=0.625,
    xtick={0,1,2,3,4,5,6,7,8,9,10,11,12,13,14},
    ytick={0.605, 0.610, 0.615, 0.62, 0.625},
    yticklabel style={/pgf/number format/.cd,fixed,precision=3},
    legend pos=outer north east,
    ymajorgrids=true,
    grid style=dashed,
    ]
    \addplot[
    color=orange,
    mark=triangle,
    ]
    coordinates {
    (1,0.613)(3,0.616)(5,0.616)(7,0.620)(10,0.616)(13,0.608)
    };
    \end{axis}
    \end{tikzpicture}
\caption{Curve for the branch selection interval $T$ on OTB2013 benchmark~\cite{otb13}.}\label{fig3} 
\end{figure}

\textbf{Multiple branches improve the tracking result.}  We compared our full branches algorithm with various combination of branches as illustrated in Table~\ref{tab1}. We evaluate the performances of the original branch, context-dependent branches and AlexNet branch alone. Note that branch selection is applied only when we evaluate the context-dependent branches, since many branches are available. For the other experiments in Table~\ref{tab1}, we combine these branches with online branch selection for testing. Results clearly demonstrate that the proposed multiple branches architecture allows a better use of diverse feature representations. The best FPS is achieved by the general siamese branch, which is expected since it needs less computations with only one branch.

\textbf{Online branch selection for every frame is not necessary.} As shown in Fig.~\ref{fig3}, we conduct experiments on the branch selection interval $T$ by changing the value: $T=1, 3, 5, 7, 10, 13$. When the value of branch selection interval is less than 7 frames, the tracking performance is reduced. This can be explained by the fact that a frequent execution of the selection mechanism increases the possibility of selecting an inappropriate branch. When the value of branch selection interval is more than 7 frames, the tracking performance is also decreased because we keep for a too long period a branch that is not discriminative anymore. In our experiments, the optimal value of branch selection interval $T$ was 7 frames.

\subsection{Comparison with State-of-the Art Trackers}
\begin{figure}
	\centering
    \subfigure{\includegraphics[width=0.3\textwidth]{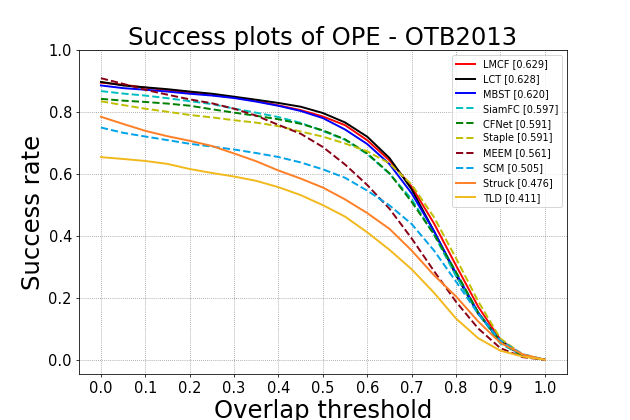}}
    \subfigure{\includegraphics[width=0.3\textwidth]{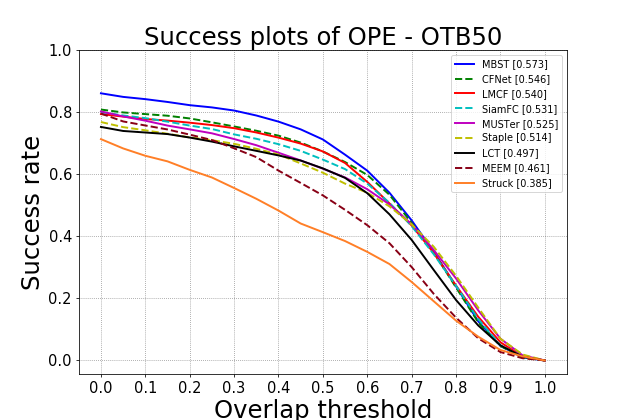}}
    \subfigure{\includegraphics[width=0.3\textwidth]{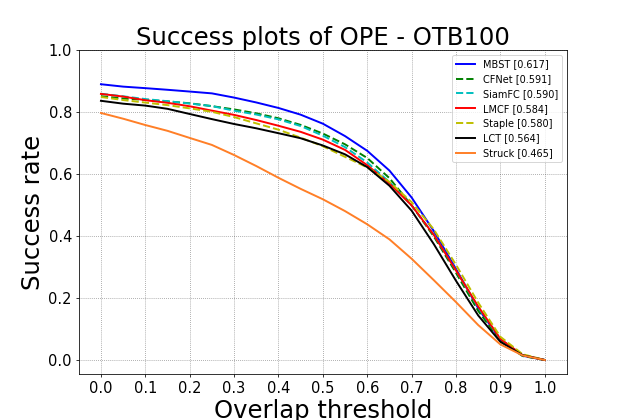}}
    \subfigure{\includegraphics[width=0.3\textwidth]{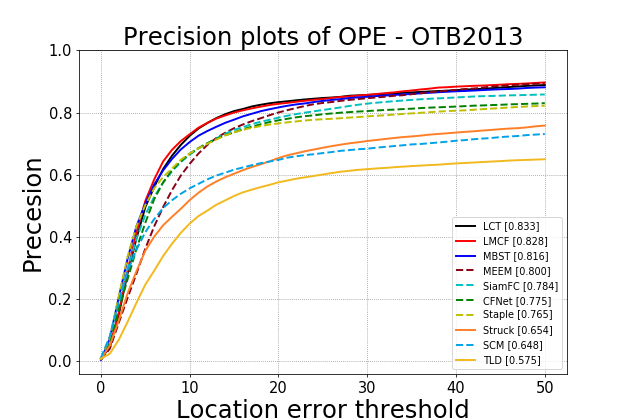}}
    \subfigure{\includegraphics[width=0.3\textwidth]{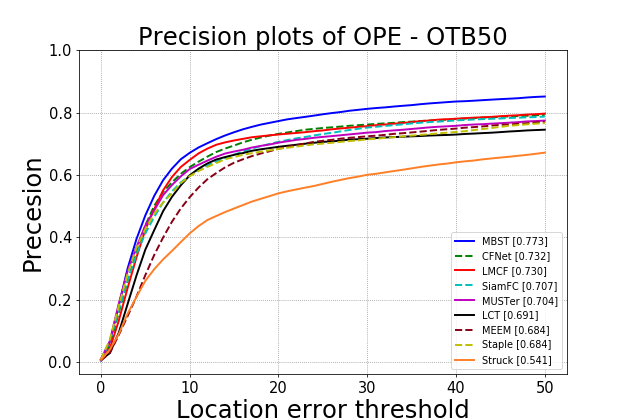}}
    \subfigure{\includegraphics[width=0.3\textwidth]{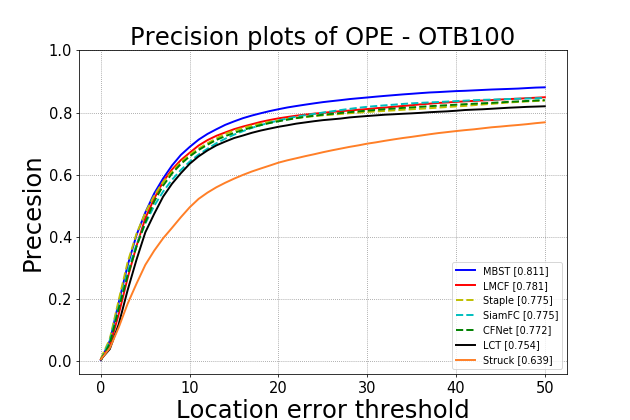}}
	\caption{The success plots and precision plots on OTB benchmarks. Curves and numbers are generated with Python implemented OTB toolkit.}\label{fig4}
\end{figure}

\begin{figure}
	\centering
    \subfigure{\includegraphics[width=0.24\textwidth]{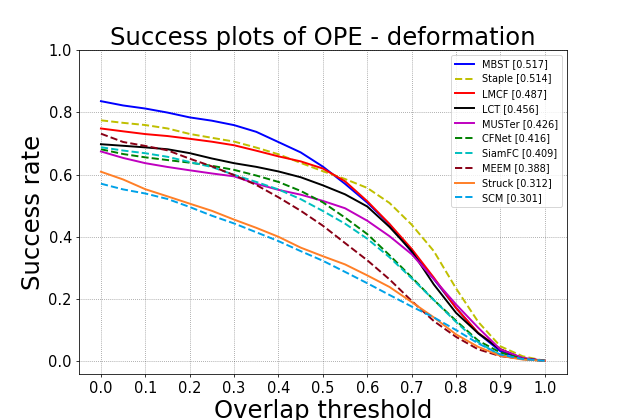}}
    \subfigure{\includegraphics[width=0.24\textwidth]{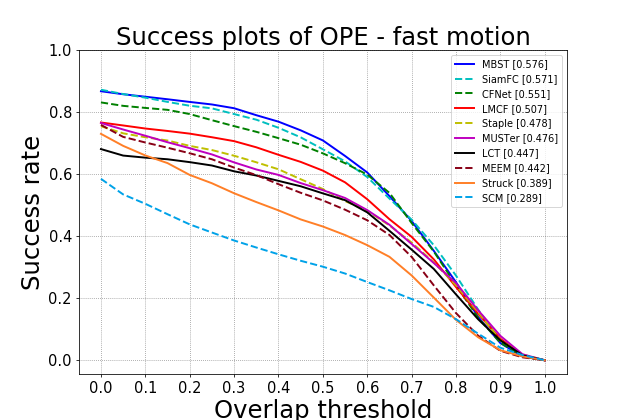}}
    \subfigure{\includegraphics[width=0.24\textwidth]{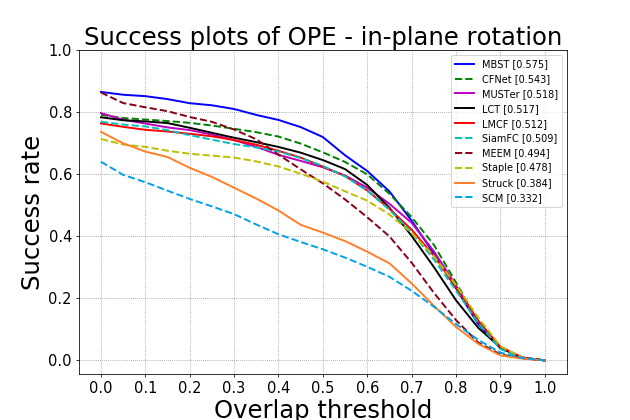}}
    \subfigure{\includegraphics[width=0.24\textwidth]{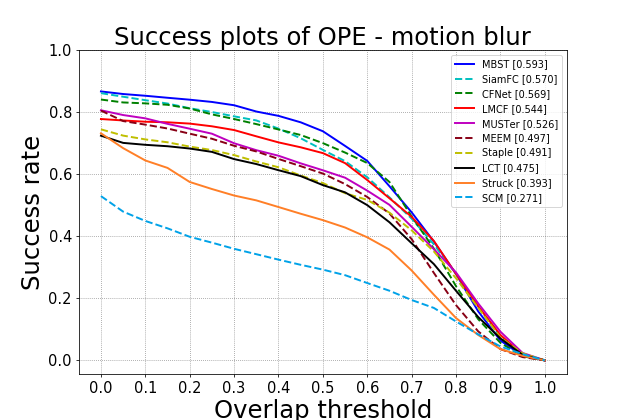}}
    \subfigure{\includegraphics[width=0.24\textwidth]{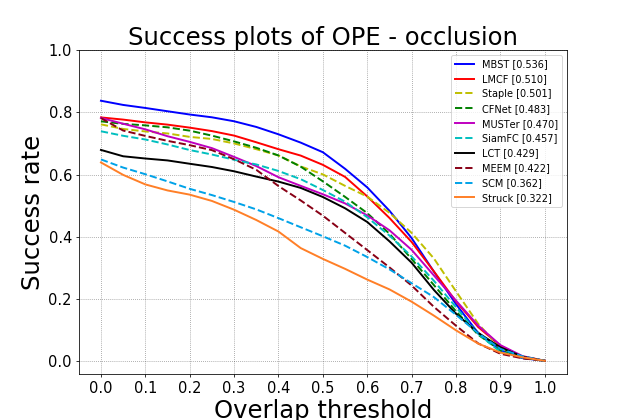}}
    \subfigure{\includegraphics[width=0.24\textwidth]{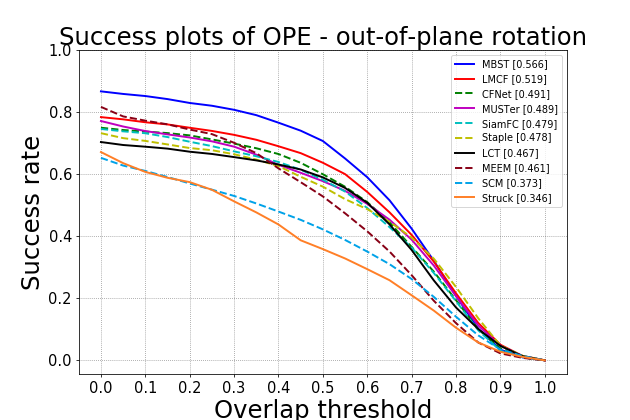}}
    \subfigure{\includegraphics[width=0.24\textwidth]{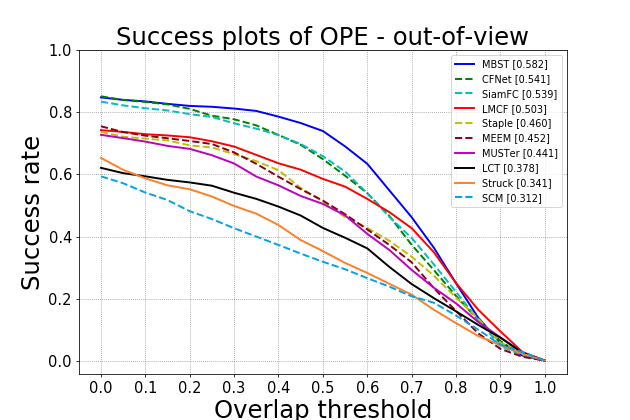}}
    \subfigure{\includegraphics[width=0.24\textwidth]{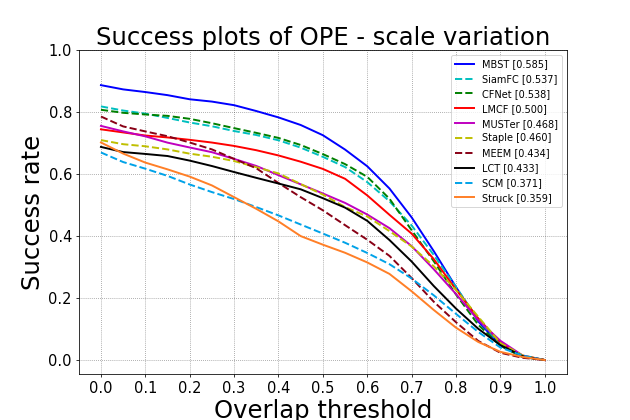}}
	\caption{The Success plot on OTB50 for eight challenge attributes: deformation, fast motion, in-plane rotation, motion blur, occlusion, out-of-plane rotation, out-of-view, scale variation.}\label{fig5}
\end{figure}

We compare MBST with CFNet~\cite{cfnet}, SiamFC~\cite{siamfc}, Staple~\cite{staple}, LCT~\cite{lct}, Struck~\cite{struck}, MEEM~\cite{meem}, SCM~\cite{scm}, LMCF~\cite{lmcf}, MUSTER~\cite{muster}, TLD~\cite{tld} on OTB benchmarks. The precision plots and success plots of one path evaluation (OPE) are shown in Fig.~\ref{fig4}. Based on precision and success plots, the overall comparison suggests that the proposed MBST achieved the best performance among these state-of-the-art trackers on OTB benchmarks. Notably, it outperforms SiamFC~\cite{siamfc} as well as its variation CFNet~\cite{cfnet} on all datasets. This demonstrates that diverse feature representations are important to improve tracking, as feature maps from various CNNs can be quite different. Fig.~\ref{fig5} demonstrates that our tracker effectively handles all kinds of challenging situations that often require high-level semantic understanding. For example, our tracker significantly outperforms SiamFC in the case of deformation, occlusion and out-of-plane rotations because the contrast between the object and the background changes and switching to another feature map may give a better discriminativity. Therefore, our approach is beneficial each time the appearance of the object changes significantly during its tracking.

\section{Conclusion}
In this paper, we propose a Multi-Branch Siamese Network with Online Selection. We ensemble multiple siamese networks to diversify target feature representations. Using our online branch selection mechanism, the most discriminative branch is selected against target appearance variations. Our tracker benefits from the diverse target representation, and can handle all kinds of challenging situations in visual object tracking. Our experiment results show improved performances compared to standard Siamese network trackers, while outperform several recent state-of-the-art trackers. 

\clearpage

%
%
%
\bibliographystyle{splncs04}
\bibliography{mybibliography}

\end{document}